\UseRawInputEncoding
\pdfoutput=1
\documentclass[letterpaper]{article} % DO NOT CHANGE THIS
\usepackage[submission]{aaai23}  % DO NOT CHANGE THIS
\usepackage{times}  % DO NOT CHANGE THIS
\usepackage{helvet}  % DO NOT CHANGE THIS
\usepackage{courier}  % DO NOT CHANGE THIS
\usepackage[hyphens]{url}  % DO NOT CHANGE THIS
\usepackage{graphicx} % DO NOT CHANGE THIS

\usepackage{amsmath,amssymb,stmaryrd}

\usepackage{natbib}  % DO NOT CHANGE THIS AND DO NOT ADD ANY OPTIONS TO IT
\usepackage{caption} % DO NOT CHANGE THIS AND DO NOT ADD ANY OPTIONS TO IT
\usepackage{multirow}
\usepackage{amssymb}
\usepackage{subcaption}
\usepackage{booktabs,tabularx}
\usepackage{algorithm}
\usepackage{algorithmic}

\usepackage{newfloat}
\usepackage{listings}
\DeclareCaptionStyle{ruled}{labelfont=normalfont,labelsep=colon,strut=off} % DO NOT CHANGE THIS
\floatstyle{ruled}
\newfloat{listing}{tb}{lst}{}
\floatname{listing}{Listing}
\title{DPTNet: A Dual-Path Transformer Architecture for Scene Text Detection}

\author{
Jingyu Lin\textsuperscript{\rm 1}\thanks{Authors contribute equally.},
Jie Jiang\textsuperscript{\rm 2}\footnotemark[\value{footnote}],
Yan Yan\textsuperscript{\rm 1},
Hanzi Wang\textsuperscript{\rm 1}\thanks{Corresponding author},
Chunchao Guo\textsuperscript{\rm 2},
Hongfa Wang\textsuperscript{\rm 2},
Wei Liu\textsuperscript{\rm 2}\\
\textsuperscript{\rm 1}Xiamen University,
\textsuperscript{\rm 2}Tencent Data Platform.\\
ljy923766649@gmail.com, \{yanyan,hanzi.wang\}@xmu.edu.cn, wl2223@columbia.edu, \{zeus,chunchaoguo,hongfawang\}@tencent.com
}

\affiliations {
    \textsuperscript{\rm 1} Affiliation 1\\
    \textsuperscript{\rm 2} Affiliation 2\\
    firstAuthor@affiliation1.com, secondAuthor@affilation2.com, thirdAuthor@affiliation1.com
}

\usepackage{bibentry}
\begin{document}

\maketitle

\begin{abstract}
The prosperity of deep learning contributes to the rapid progress in scene text detection. Among all the methods with convolutional networks, segmentation-based ones have drawn extensive attention due to their superiority in detecting text instances of arbitrary shapes and extreme aspect ratios. However, the bottom-up methods are limited to the performance of their segmentation models. In this paper, we propose DPTNet (Dual-Path Transformer Network), a simple yet effective architecture to model the global and local information for the scene text detection task. We further propose a parallel design that integrates the convolutional network with a powerful self-attention mechanism to provide complementary clues between the attention path and convolutional path. Moreover, a bi-directional interaction module across the two paths is developed to provide complementary clues in the channel and spatial dimensions. We also upgrade the concentration operation by adding an extra multi-head attention layer to it. Our DPTNet achieves state-of-the-art results on the MSRA-TD500 dataset, and provides competitive results on other standard benchmarks in terms of both detection accuracy and speed.
\end{abstract}

% a bi-directional interaction module across the two paths

\section{Introduction}

In the past few years, scene text processing has attracted a surge of research interest and plays an important role in wide-range applications such as sign board recognition, autonomous driving, and scene parsing. Scene text detection aims to detect and localize text regions in natural images. Different from object detection, scene text detection mainly focuses on the detection of various irregularly shaped texts in natural scenes. When it comes to the natural scenario, the unique traits of texts in geometric layouts (e.g., large aspect ratios, various scales, random rotations, and curve shapes) remain a gap for researchers to get better performance.

\begin{figure}[tbp]
\centering
\includegraphics[width=0.9\linewidth]{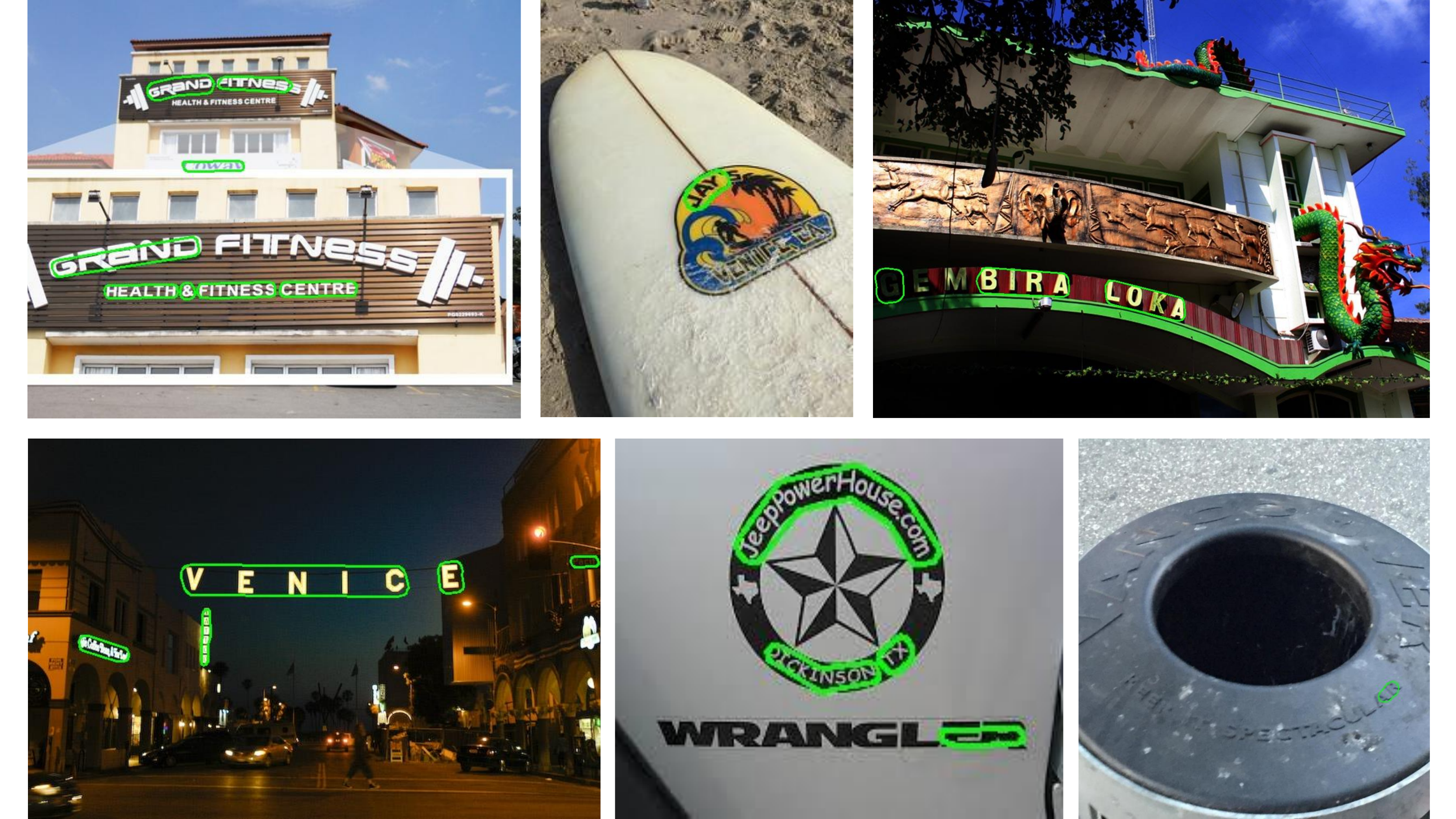}
\caption{Some detection results of text instances in natural scenes, generated by the classical bottom-up method.}
\label{badcase}
\end{figure}

Since a large portion of text instances in the natural scene is usually distorted with a peculiar shape and size, the main challenge for scene text detection is to localize the region of each text instance correctly with integrity. Recently, segmentation-based scene text detection has attracted much attention, and these methods \cite{ZhengZhang2016MultiOrientedTD,ChuhuiXue2018AccurateST,ZhuotaoTian2019LearningSE,WenhaiWang2019ShapeRT,ZhuotaoTian2019LearningSE} make the detection work a two-stage task that makes full use of pixel-level predictions and post-processing algorithms. From the pixel point of view, a text region can be regarded as the image classification task if we give per-pixel binary prediction (text or non-text) instead of image-level bounding box prediction. Therefore, the performance of binary class segmentation in this problem has become the bottleneck to deal with the problem itself. Moreover, they may fail when applied to scene text images due to existing extreme aspect ratio issues of natural text instances, such as the extreme long text instances and wide spacing instances, etc (see Figure \ref{badcase} for some examples). To better understand the semantics of an image, context information is usually essential for a certain scene text detection model. 

To enlarge the receptive field, Transformer \cite{AshishVaswani2017AttentionIA} turns out to be a good choice. Dosovitskiy et al. proposed Vision Transformer(ViT) \cite{AlexeyDosovitskiy2020AnII} to split the image into patches which are treated as sequence features and then fed to a standard Transformer with positional embeddings to generate a global interaction between different components, leading to an excellent performance on the ImageNet dataset. The well-designed architectures \cite{HugoTouvron2020TrainingDI,NicolasCarion2020EndtoEndOD,WenhaiWang2021PyramidVT,HaipingWu2021CvTIC} are either working individually or becoming an excellent auxiliary part for CNNs in various visual tasks such as generation, detection, segmentation, etc. However, self-attention in Transformer blocks is not a panacea, where researchers found that a loss of local details is easily ignored in the module. For scene text detection, there are also some DETR-based frameworks \cite{NicolasCarion2020EndtoEndOD} which bring advantages in global-range feature modeling but which may not handle the target with a typical size. And some other works \cite{DepuMeng2021ConditionalDF,XizhouZhu2020DeformableDD} try to improve the efficiency of transformer-based object detectors by optimizing the attention operations, but the results they achieve in scene text detection are not that competitive. For example, the performance of a recent DETR-based scene text detector \cite{ZobeirRaisi2021TransformerbasedTD} deteriorates when it comes to text instances that have much larger variances of scales and aspect ratios. It turns out that pure Transformers cannot obtain sufficient information by themselves.

In this paper, we propose our DPTNet (Dual-Path Transformer Architecture) for the scene text detection task which, jointly considers the attention at the whole scale and local feature aggregation. The novelties of our approach are three-fold:

\begin{itemize}
\item We propose a dual-path hierarchical backbone to empower CNN with global attention, significantly improving the text detection performance.
\item We propose a simple yet effective design named bi-directional connection module that enables the context information communication between two parallel paths. And the entire model yields scale-robust features by a multi-head attention block at the feature concentrating stage without invoking complex and computationally heavy modules.
\item Using the proposed method, we achieve competitive results in terms of efficiency, accuracy, and robustness on four public natural scene text detection datasets.
\end{itemize}

\section{Related Work}
With the booming development of deep learning, traditional heuristic methods in scene text detection have been gradually replaced by neural-network-based methods, which show brilliant performance in practical applications. 
% \subsection{Scene Text Detection}

{\bfseries Scene Text Detection.} Traditional object detection methods are also applicable to text detection. These methods \cite{liao2017textboxes,liao2018textboxes++,liu2017deep,zhou2017east,WenhaoHe2017DeepDR,liao2018rotation} often take advantage of the classical object detector to directly predict bounding box coordinates of text instances in an end-to-end manner based on SSD \cite{liu2016ssd} or inspired by FCN \cite{dai2016r}. However, most of them cannot well represent accurate bounding boxes for irregular shapes, such as curved shapes. To deal with long instances, SegLink \cite{BaoguangShi2017DetectingOT} and SegLink++ \cite{JunTang2019SegLinkDD} predict all segments within a text line and then learn the relation between segments to form the bounding boxes. Although these methods try to find new ways to deal with the post-processing procedure, they suffer from a greater computational complexity and a lower inference speed. 

% \subsection{Convolution and Global Attention}

{\bfseries Global and Local Clues.}
CNNs build a hierarchical framework to aggregate the regional information stage by stage with kernels of fixed size and have been proven to be powerful for their ability in feature extraction. A major factor limiting the performance of CNNs is the size of the receptive field. Most CNNs \cite{AlexKrizhevsky2012ImageNetCW,KarenSimonyan2014VeryDC,KaimingHe2015DeepRL,ChristianSzegedy2016Inceptionv4IA,SainingXie2016AggregatedRT} often focus on local features and inevitably ignore the context of an image due to its structural characteristics. To alleviate the above limitation, researchers have made many attempts to optimize the receptive fields over the years. The global attention mechanism \cite{XiaolongWang2017NonlocalNN,YueCao2019GCNetNN,IrwanBello2019AttentionAC,AravindSrinivas2021BottleneckTF} was used to capture long-distance dependencies in the natural language processing \cite{AshishVaswani2017AttentionIA,JacobDevlin2018BERTPO} field at first. The attention methods \cite{JieHu2022SqueezeandExcitationN,SanghyunWoo2018CBAMCB} use the deeper architecture and global average pooling operation to offer a channel-wise and spatial weight along the feature channels. A dilated convolution kernel \cite{FisherYu2015MultiScaleCA} was proposed to increase the sampling step size. Deformable convolution \cite{JifengDai2017DeformableCN} applied adaptive offsets to the convolution kernels, making the shape of the receptive field adaptive.

In total, prevailing regression-based methods need to regress a fixed number of coordinate points to form a complete text area, which has obvious shortcomings in the detection of irregular-shaped text in natural scenes. In practical applications, scene text detection is often processed offline as a pre-task of text recognition and the segmentation-based method has better performance in the offline scenario. And our method also provides a light model named DPTNet-Tiny that achieves real-time detection with little impact on accuracy, which is not inferior to the one-stage regression-based method. 

% Our dual path network explores a efficient way for interactions across attention branches and CNN branch, it is light-weighted and improve the modeling ability without bells and whistles. 

\section{Methodology}

\begin{figure*}[tbp]
\centering
\includegraphics[width=0.9\linewidth]{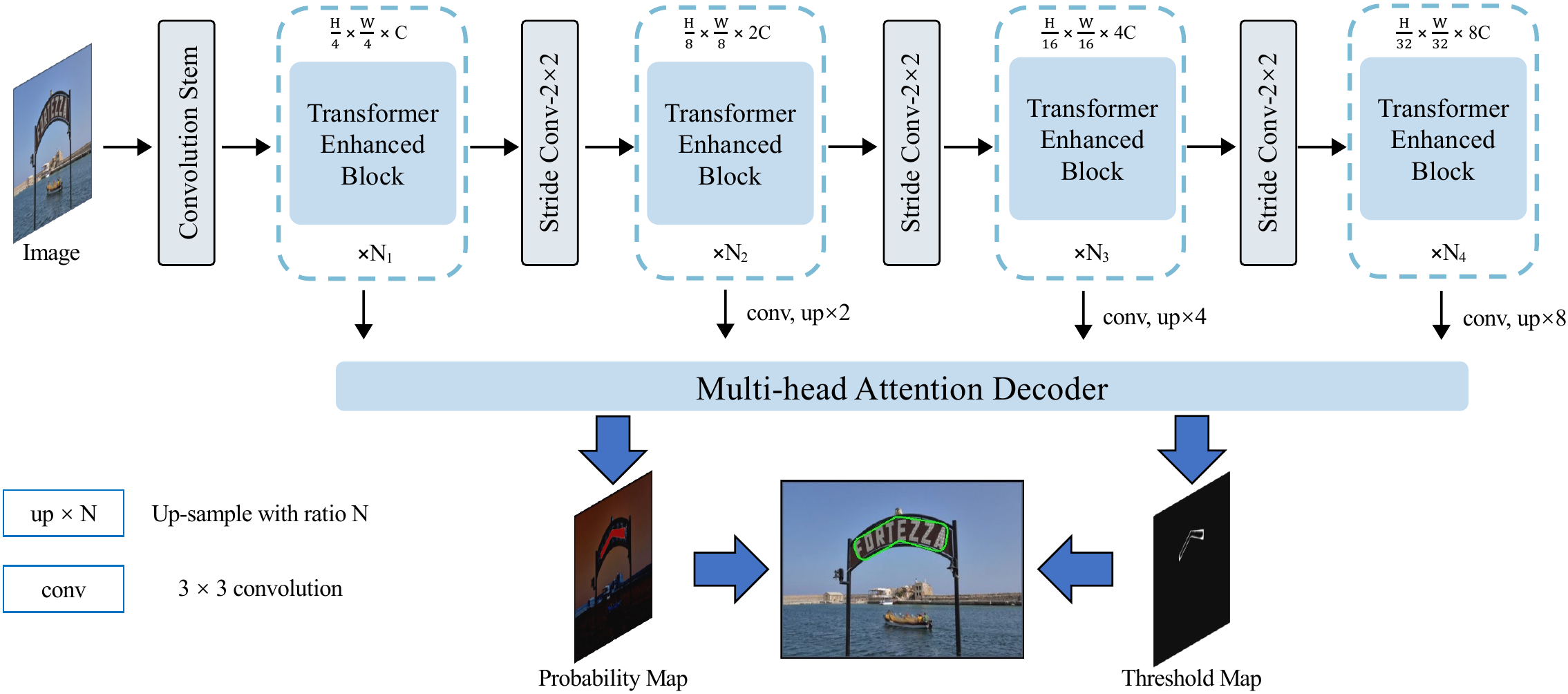}
\caption{\textbf{The Overall Architecture of DPTNet.} A feature-pyramid backbone to extract coarse and fine features of different scales; a lightweight multi-head attention decoder to fuse these multi-level features, and a training strategy to include a differentiable binarization procedure for generating a better binary map.}
\label{pipeline}
\end{figure*}

% \subsection{Architecture}
The architecture of our proposed DPTNet model is shown in Figure \ref{pipeline}. The pipeline can be divided into three parts: a Transformer-enhanced backbone, a multi-head attention decoder head, and a post-processing procedure. Firstly, similar to the traditional CNN backbone, the input image is fed to an FPN structure \cite{TsungYiLin2016FeaturePN} in order to obtain multi-level features and there are four stages with downsampling rates of \{4, 8, 16, 32\}, respectively. Secondly, we up-sample the extracted features to the same scale, and apply multi-head attention to extract the contextual feature. The feature will then be used to generate the probability map and the threshold map.

\subsection{Transformer-enhanced Backbone}

Since both CNN and Transformer have specific limitations but complementary advantages, we naturally make them work parallel for gaining better performance.

\begin{figure}[htbp]
\centering
\includegraphics[width=1.0\linewidth]{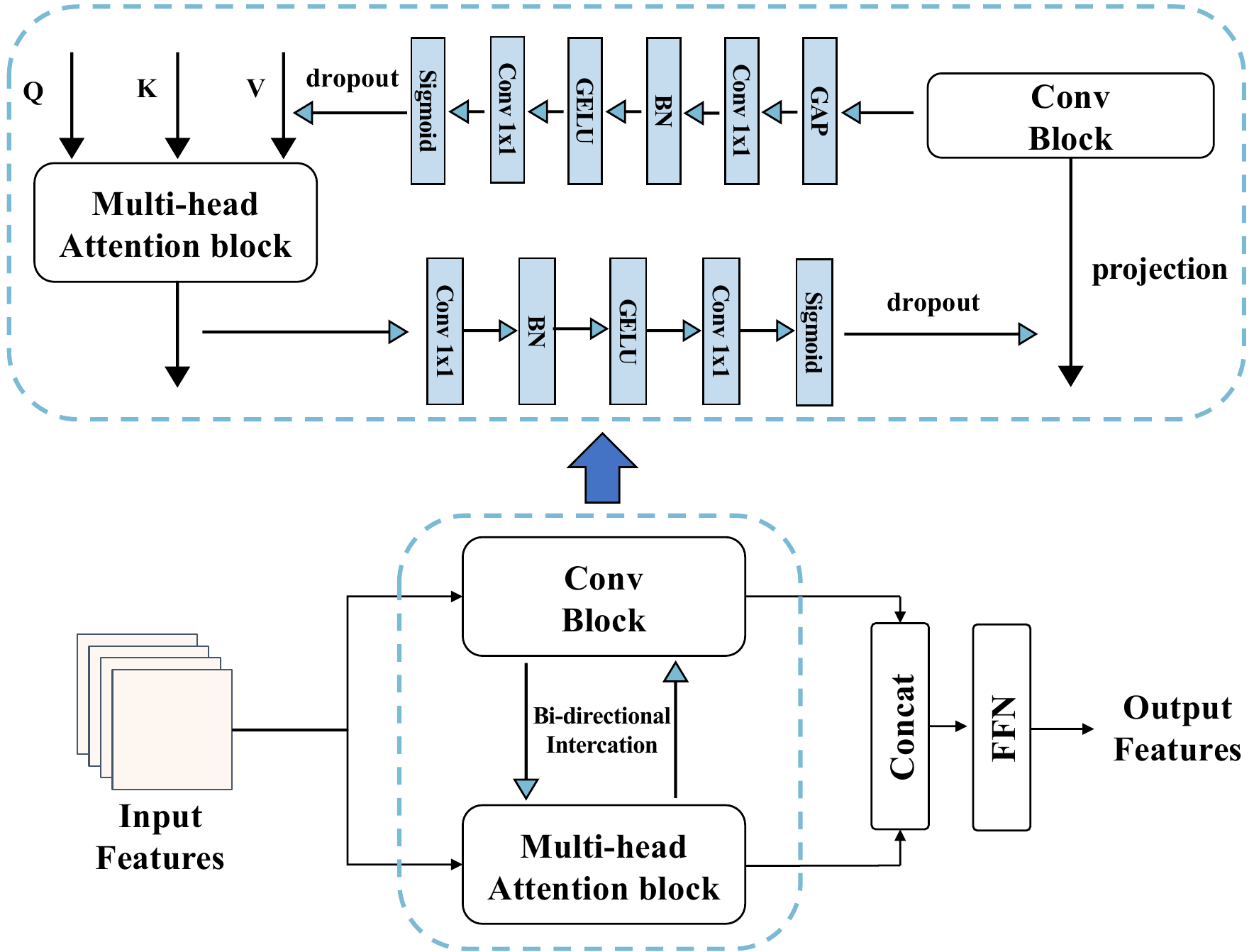}
\caption{\textbf{The Hybrid Transformer Block.} We combine self-attention with convolution in a dual-path way. It aims to capture the information within and across original receptive fields by dealing with input features in two parallel branches, transmitting messages to the Feed-Forward Network(FFN) for outputting features.}
\label{mixing}
\end{figure}

{\bfseries Parallel Design.} As we know, the inductive bias of convolutional layers endows itself with the ability to model local relations. As illustrated in Figure \ref{mixing}, a \textit{$4 \times 4$} window is adopted in local-window self-attention, similar to previous works \cite{EnzeXie2021SegFormerSA, ZeLiu2021SwinTH}. While in the CNN path, we apply depth-wise convolution with a \textit{$3 \times 3$} kernel size for efficiency. Moreover, due to the architectural difference between the two paths, we need to adjust the number of channels to match the merged branch such that they will be integrated smoothly. After the channel adjustment, their outputs are normalized by different normalization layers \cite{SergeyIoffe2015BatchNA} and concentrated together. During the training procedure, the two parallel branches can be optimized simultaneously, so that the interweaving features across the two branches will achieve stronger feature representation learning. This is followed by a successive Feed-Forward Network (FFN) to fuse the learned relations in both paths, generating the final output features.

{\bfseries Feature Interaction Module.} 
As we can see in Figure \ref{mixing}, the output feature of dual-path branches is not simply concentrated, and there exist bi-directional interactions between them. The parallel branches provide complementary clues for better representation learning in both branches. The channel/spatial interaction provides the channel/spatial context extracted by depth-wise convolution/local-window self-attention to the other path. The bi-directional interactions in the same block consist of the channel and spatial interaction. Firstly, we apply a design similar to the SE layer \cite{JieHu2022SqueezeandExcitationN}, such that the information in the convolution branch flows to the other branch through the channel interaction, which contains one GAP(global average pooling) layer, two \textit{$1 \times 1$} convolutional layers with normalization \cite{SergeyIoffe2015BatchNA} and activation \cite{DanHendrycks2016GaussianEL} between them, and a sigmoid to offer a dynamic weight to different channels. Secondly, two successive \textit{$1 \times 1$} convolutional layers with BN \cite{SergeyIoffe2015BatchNA} and GELU \cite{DanHendrycks2016GaussianEL} between them reduce the channel number to 1 with information from the self-attention branch. Also, a sigmoid layer for spatial weight distribution is used. 

% The channel/spatial interaction provides the channel/spatial context extracted by depth-wise convolution/local-window self-attention to the other path. The bi-directional interactions in the same block consist of the channel and spatial interaction.

% \begin{equation}
% \hat{X_{i}} = {\rm Concat}({\rm LayerNorm}(X_{i}), {\rm MSA}(X_{i}), {\rm CONV}(X_{i}) + X_{i})
% \end{equation}

{\bfseries Hybrid Transformer Block.} 
Our hybrid Transformer block consists of a parallel design with efficient operations, a feature interaction module, and an FFN network \cite{AshishVaswani2017AttentionIA}. Residual connections are also introduced, formulated as follows:
\begin{equation}
\begin{aligned}
% \begin{split}
    % \setlength\abovedisplayskip{0.1cm}
    % \setlength\belowdisplayskip{0.1cm}
    \hat{X_{i}} = {\rm Concat}( & {\rm LayerNorm}(X_{i}), {\rm MSA}(X_{i}),\\
    & {\rm Conv}(X_{i}) + X_{i}),
    % \hat{F}_{i} = \{&UpSample(Conv(P_{i-1})),Conv(P_{i}),\\
    % &DownSample(Conv(P_{i+1}))  \}
% \end{split}
\end{aligned}
\end{equation}
\begin{equation}
X_{i+1} = {\rm FFN}({\rm LayerNorm}(\hat{X_{i}})) + \hat{X_{i}},
\end{equation}
where \textit{$X_{i}$} stands for the features at the i-th block, MSA means the multi-head self-attention branch, and Conv is a symbol of the depth-wise convolution operation. For FFN, we follow previous works \cite{ZeLiu2021SwinTH,HugoTouvron2020TrainingDI} to set an MLP that consists of two linear layers with one GELU \cite{DanHendrycks2016GaussianEL} between them.

% and send features to the Feed-Forward Network(FFN) for outputting features.
% It aims to capture the information within and across original receptive fields by dealing with input features in two parallel branches, transmitting messages to the Feed-Forward Network(FFN) for outputting features.

\subsection{Multi-head Attention Decoder}
Most semantic segmentation methods use cascading or summing up methods to fuse the features from different scales. However, such a simple fusion paradigm ignores attention and loses some essential details inevitably. Therefore, we use a multi-head attention decoder to dynamically re-attend to the instances and retain the integrity of the text region hidden at the spatial level. 

We scale the features from different levels into the same resolution first so that they can be processed at the same scale. We denote the feature maps generated from the four stages in our backbone as \textit{$X \in R^{N \times C \times H \times W} =  \left \{ X_{i}  \right \}_{i=0}^{N-1} $}, where N is equal to 4. We use a \textit{$3 \times 3$} convolutional layer to compress the channels to a fixed smaller number and get an intermediate feature \textit{$S \in R^{C \times H \times  W}$}. Then a simple but effective multi-head attention module splits the features into $N$ parts (N is 4 here) along the channel dimension. The data-dependent dynamic weights will be assigned to the corresponding scaled features to get the final fused feature F. The above operations can be formulated as:
\begin{equation}
    \hat{F} = {\rm Conv}({\rm Concat}(\left [ X_{0}, X_{1} … , X_{N-1}   \right ] )),
\end{equation}
\begin{equation}
    F = {\rm MSA}(\hat{F} ).
\end{equation}
Here \textit{${\rm Concat}(\cdot)$} indicates the concatenation operator; \textit{${\rm Conv}(\cdot)$} is the \textit{$3 \times 3$} convolutional layer. MSA is short for Multi-head Self-attention. We set the head number the same as the stage number to 4. Here $i$ stands for the index of the head, \textit{$i \in \left \{ 0, 1, 2, 3\right \}$}.

\subsection{Post-processing}

For the obtained features, appropriate post-processing strategies can make the text regions we get more expressive. In the process of parsing the features into the text regions we visualize, we need to perform binarization and label generation operations over the features.

{\bfseries Differentiable Binarization.}
The differentiable binarization for the probability map was first used by \cite{MinghuiLiao2019RealtimeST}, which provided a trainable binarization method to generate a segmentation mask. The output of the segmentation network is a probability map \textit{$P \in R^{H \times W}$}, where \textit{H} and \textit{W} stand for the height and width of the map. For a text-or-not judging task, a binary map \textit{$B \in R^{H \times W}$} is needed, where the pixels with the value 1 are considered as valid text areas.  Usually, the standard binarization process can be formulated as follows:
\begin{equation}
B_{i, j}=\left\{\begin{array}{ll}
1 & \text { if } P_{i, j}>=t,  \\
0 & \text { otherwise }.
\end{array}\right.
\end{equation}
Here \textit{t} is a predefined threshold number, and (\textit{i}, \textit{j}) refers to the value of pixel (\textit{x}, \textit{y}) in the probability map. In contrast, differentiable binarization is especially beneficial to distinguishing text regions from the background and separating text instances which are closely jointed. Therefore, an approximate step function is proposed as follows:
\begin{equation}
\hat{B}_{i,j} = \frac{1}{1 + e^{-k(P_{i,j} - T_{i,j})} }.
\end{equation}
The above \textit{$\hat{B}$} is the approximate binary map; \textit{T} is the adaptive threshold map learned from the network. \textit{k} is a hyperparameter set as 50 empirically to behave similar to the standard binarization but is differentiable so that it can be optimized along with the segmentation network in the training period.

{\bfseries Adaptive Threshold. } An adaptive threshold was firstly proposed in \cite{ChuhuiXue2018AccurateST} in the form of the so-called text border map and upgraded by \cite{MinghuiLiao2019RealtimeST} which applied border-like supervision on the threshold map for better guidance. Each pixel in the text border map here is served as the threshold according to the probability map for the binarization.

\subsection{Label Generation}

The label generation for the probability map is inspired by PSENet \cite{WenhaiWang2019ShapeRT}. Usually, post-processing algorithms present the segmentation results by a group of vertexes that forms a polygon:
\begin{equation}
    G = \left \{ S_{k}  \right \} _{k=1}^{n}. 
\end{equation}

% To solve the problem that the boundaries of adjacent texts are difficult to define, 
% Vatti clipping algorithm \cite{BalaRVatti1992AGS} is proved to be effective in generating an offset for shrinking the original polygon. 

$n$ is the number of vertexes, which usually differs according to the labeling rule in various datasets, and \textit{$S$} is a symbol for segmentation results in each image. To solve the problem that the boundaries of adjacent texts are difficult to define, Vatti clipping algorithm \cite{BalaRVatti1992AGS} was proposed to effectively generate an offset for shrinking the original polygon. The offset D can be mathematically calculated as:
\begin{equation}
    D = \frac{{\rm Area}(P) \times (1 - r^{2} )}{{\rm Perimeter}(P)}. 
\end{equation}
Here \textit{$Area(\cdot)$} means the computing of the polygon area, similarly, \textit{$Perimeter(\cdot)$} means the computing of the polygon perimeter, and $r$ is the shrink ratio, which is set to 0.4 empirically. Using graphics-related operations, the results of shrunk polygons can be easily obtained from the original ground truth, and become the kernel for each text region. 

% In addition to the shrinking operation, we also need to restore the text area to the size of the original label for the threshold map. The outward offset of the expansion operation is consistent with the \textit{$D$} value in the above formula. We denote the shrunk kernel by \textit{$G_{k}$} and the expansion region boundary by \textit{$G_{e}$}, respectively. The values between \textit{$G_{k}$} and \textit{$G_{e}$} are larger in the middle and smaller on both sides. The pixels on the boundary of \textit{$G_{k}$} and \textit{$G_{e}$} are the lowest, and the values on the contour line of the original ground truth are the highest.

\begin{figure}[!htbp]
\centering
\includegraphics[width=1.0\linewidth]{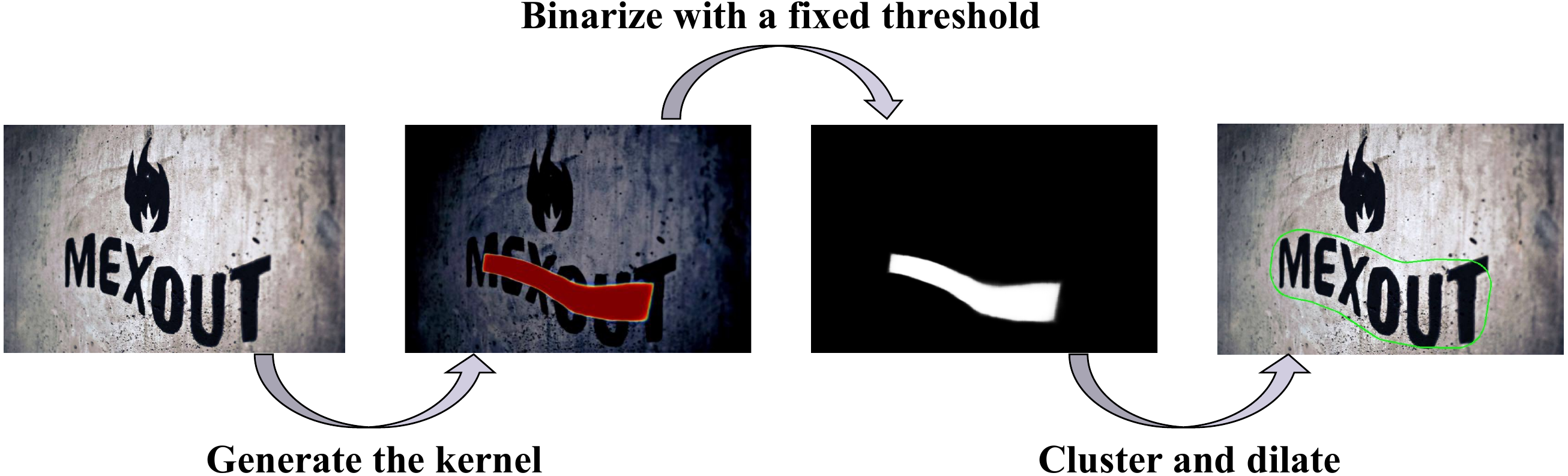}
\caption{\textbf{The Inference Process.} We use the probability map so that the threshold branch can be removed for better efficiency.}
\label{dilate}
\end{figure}

To simplify the inference process, we can only use the probability map to obtain the box formation as shown in Figure \ref{dilate}. Firstly, the probability map will be binarized with a fixed threshold (usually set as 0.2 here) to get the binary map. Secondly, we will cluster the text pixels to obtain shrunk text regions according to the binary map. The final text prediction results are dilated from shrunk regions with an offset \textit{$D{}'$} as:
\begin{equation}
    D{}' = \frac{{\rm Area}(P) \times r{}'}{{\rm Perimeter}(P)}. 
\end{equation}
Here $r{}'$ is set to 1.5 empirically.

\subsection{Optimization}
For training the DPTNet, the loss function can be formulated as:
\begin{equation}
    L = L_{s} + \alpha \times L_{b} + \beta \times L_{t}. 
\end{equation}
We assign different weights to the probability map loss \textit{$L_{s}$}, the binary map loss \textit{$L_{b}$}, and the threshold map \textit{$L_{t}$} accordingly. In practice, we set \textit{$\alpha$} as 1.0 and \textit{$\beta$} as 10, respectively. For \textit{$L_{s}$} and \textit{$L_{b}$}, the loss function used here is binary cross-entropy (BCE) loss. Aiming to overcome the imbalance of the numbers of positives and negatives, which usually occurs in segmentation-based text detection methods, hard negative mining is applied in the BCE loss by sampling the hard negatives and formulated as follows:
\begin{equation}
    L_{s} = L_{b} = \sum_{i \in S_{l}}^{} y_{i}log{x_{i}} + (1 - y_{i})log(1 - x_{i}),
\end{equation}
where \textit{$S_{l}$} is the set sampled by a ratio of positives and negatives at 1 : 3, which is called Online Hard Example Mining (OHEM) \cite{AbhinavShrivastava2016TrainingRO}. To calculate \textit{$L_{t}$}, we use $L_{1}$ distances to measure the similarity between the prediction and label inside the dilated text polygon:
\begin{equation}
    L_{t} = \sum_{i \in R_{d}}^{} \left | y_{i}^{*} - x_{i}^{*}   \right |.
\end{equation}
Here, \textit{$R_{d}$} is a set of pixels inside the dilated polygon; \textit{$y^{*}$} is the label for the threshold map, and \textit{$x^{*}$} is the prediction probability map.

% which is made up of all the positives and top-k negatives with k set as 3 times the number of positives. 

\begin{table*}[!ht]
\centering
\caption{Detection results with different settings of the proposed bi-directional connection module, Transformer-enhanced backbone. \textit{BC} indicates the bi-directional connection module. \textit{TEB} indicates the Transformer-enhanced backbone. \textit{P}, \textit{R}, \textit{F} indicate precision, recall, and F1-score, respectively.}
\begin{tabularx}{1.0\linewidth}{@{}l*{10}X@{}}
\toprule
% \begin{tabular}{|c|c|c|c|c|c|c|c|c|c|c|}
% 
\multirow{2}{*}{Backbone} & \multirow{2}{*}{BC} & \multirow{2}{*}{TEB} & \multicolumn{4}{c}{MSRA-TD500} & \multicolumn{4}{c}{CTW1500} \\ \cline{4-11} 
    &    &    & P      & R      & F      & FPS & P      & R      & F      & FPS \\ 
\midrule   
ResNet-18   & $\times$   & $\times$   & \textbf{90.4}   & 76.3   & 82.8   & \textbf{62} & 84.8 & 77.5 & 81.0 & \textbf{55} \\ 
ResNet-18   & \checkmark  & $\times$ & 88.7   & 77.9   & 82.9   & 50  & 84.7 & 78.5 & 81.4 & 42 \\ 
DPTNet-tiny(ours)   & $\times$   & \checkmark & 88.8   & 81.1   & 84.8   & 52  & 86.1 & 80.9 & 83.4 & 44 \\ 
DPTNet-tiny(ours)   & \checkmark  & \checkmark  & 88.4   & \textbf{82.9}   & \textbf{85.6}   & 49  & \textbf{86.4} & \textbf{81.5} & \textbf{84.1} & 43  \\ 

\midrule  

ResNet-50  & $\times$   & $\times$  & 91.5   & 79.2   & 84.9   & \textbf{32}  & 86.9 & 80.2 & 83.4 & \textbf{22} \\ 
ResNet-50  & \checkmark & $\times$   & 90.2   & 80.4   & 85.0   & 21 & 86.3 & 81.1 & 83.6 & 16 \\ 
DPTNet-normal(ours) & $\times$ & \checkmark  & 92.3   & 83.7   & 87.7   & 27 & 87.4 & 82.8 & 85.0 & 19 \\ 
DPTNet-normal(ours) & \checkmark & \checkmark & \textbf{92.5}  & \textbf{84.3}    & \textbf{88.2}   & 26  & \textbf{88.1} & \textbf{83.1} & \textbf{85.5} & 19 \\ 
\bottomrule
\end{tabularx}
\label{teb}
\end{table*}

\section{Experiments}

\subsection{Datasets} \label{data}

We use five standard arbitrary-shaped scene text detection datasets to evaluate the performance of DPTNet. Apart from the large pre-training dataset Synthtext \cite{AnkushGupta2016SyntheticDF}, we divide the datasets into 3 types as follows:

{\bfseries Curved Text Detection.} 
Total-Text \cite{CheeKhengChng2017TotalTextAC} is a dataset consisting of 1255 training images and 300 testing images with texts of various shapes, including horizontal, multi-oriented, and curved ones. We also test the performance of our method on CTW-1500 \cite{YuliangLiu2019CurvedST}. A challenging dataset focuses on long curve text detection. It consists of 10751 text instances in 1000 training images and 500 testing images. Every image is annotated at the text-line level by a region of 14 vertices.

{\bfseries Multi-Oriented Text Detection.}
ICDAR2015 \cite{DimosthenisKaratzas2015ICDAR2C} is a commonly used dataset for text detection. It was firstly introduced in the ICDAR 2015 Robust Reading Competition for incidental scene text detection. The images are captured by Google glasses with a resolution of 720 × 1280. It contains 1000 images with labels for training and 500 for testing. The annotations are at the word level and the locations of texts are labeled with quadrilateral boxes.

% MLT-2017 \cite{ic17} is a large scale multi-lingual text dataset. It contains 7,200 training images, 1,800 validation images, and 9,000 test images. The dataset is composed of complete scene images which come from 9 languages representing 6 different scripts. The text regions are annotated by the 4 vertices of quadrilaterals. 

{\bfseries Multi-Language Text Detection.}
We apply our method to a smaller dataset named MSRA-TD500 \cite{CongYao2012DetectingTO}. It is a multi-language dataset that includes English and Chinese. It consists of 300 training and 200 testing images. The annotations are in the word-line level, whose text regions are labeled with rotated rectangles.

% Since SynthText provides large quantities of synthetic words so that it offer better prior knowledge for real-world datasets. 

\subsection{Implementation Details} \label{detail}

Before training on the real-world datasets, we perform the pre-training work on the SynthText dataset for 160k iterations. After the pre-training, we finetune the models for 1200 epochs on the corresponding real-world datasets. Our data augmentation for training mainly includes random rotation, random cropping, random horizontal, and vertical flipping. And we resize all the images to 640 $\times$ 640 for better training efficiency. We set the training batch size to 16 for all datasets and follow a "poly" learning rate policy to make the learning rate decay gradually. We set the initial learning rate to 0.007 with the attenuation coefficient of 0.9. Stochastic gradient descent (SGD) is adopted to optimize our framework, and the weight decay and the momentum are set to 0.0001 and 0.9, respectively.

% By comparing the two variables in a controlled experiment, we can evaluate how each module influences final performance and verify the effectiveness. 

\begin{table*}[!tb]
    \centering
    \caption{Detection results on ICDAR2015, MSRA-TD500, Total-Text, and CTW1500. P, R, and F represent Precision, Recall, and F-score (in \%), respectively. The best results in each column are highlighted by bold.}
    \begin{tabularx}{1.0\linewidth}{@{}l*{12}X@{}}
    % {lcccccccccccc}
    \toprule
    \multirow{2}{*}{Method} & \multicolumn{3}{c}{ICDAR2015} & \multicolumn{3}{c}{MSRA-TD500} & \multicolumn{3}{c}{Total-Text} & \multicolumn{3}{c}{CTW1500}\\  \cline{2-13}
    ~ & P & R & F & P & R & F& P & R & F& P & R & F \\ \midrule
    % EAST~\cite{EAST}& 83.6  & 73.5  & 78.2  & 87.3  & 67.4  & 76.1 &-&-&- &-&-&-\\
    % DDR~\cite{DDR}  & 82.0  & 80.0  & 81.0 &77.0 &70.0 &74.0 &-&-&- &-&-&-\\
    % Corner~\cite{Corner} & 94.1  & 70.7  & 80.7& 87.6  & 76.2  & 81.5 &-&-&- &-&-&-\\
    % TLOC~\cite{ctw1500-pr} &-&-&-& 84.5  & 77.1  & 80.6  &74.0&71.0&72.5 & 77.4  & 69.8  & 73.4 \\
    TextSnake~\cite{ShangbangLong2018TextSnakeAF} & 84.9  & 80.4  & 82.6 & 83.2  & 73.9  & 78.3 & 82.7  & 74.5  & 78.4  & 67.9  & 85.3  & 75.6 \\
    TextField~\cite{TextField} & 84.3 & 83.9 & 84.1 & 87.4 & 75.9 & 81.3  & 81.2          & 79.9          & 80.6  & 83.0          & 79.8          & 81.4  \\
    PSE-Net~\cite{WenhaiWang2019ShapeRT} & 86.9 & 84.5 & 85.7  &-&-&-& 84.0          & 78.0          & 80.9   & 84.8          & 79.7          & 82.2   \\
    LOMO~\cite{ChangqianYu2018LearningAD} & 91.3  & 83.5  & 87.2 &-&-&-& 88.6 & 75.7  & 81.6  & \textbf{89.2} & 69.6 & 78.4\\
    % ATRR~\cite{ATRR}    & 89.2 & 86.0 & 87.6  & 85.2 & 82.1 & 83.6  & 80.9 & 76.2  & 78.5  & 80.1 & 80.2 & 80.1\\
    CRAFT~\cite{YoungMinBaek2019CharacterRA} & 89.8 & 84.3 & 86.9 & 88.2 & 78.2 & 82.9    & 87.6          & 79.9          & 83.6  & 86.0          & 81.1          & 83.5   \\
    PAN~\cite{WenhaiWang2019EfficientAA} & 84.0 & 81.9 & 82.9 & 84.4 & 83.8 & 84.1   & 89.3          & 81.0          & 85.0 & 86.4          & 81.2          & 83.7  \\
    DB~\cite{MinghuiLiao2019RealtimeST} & 91.8 & 83.2 & 87.3  & 91.5 & 79.2 & 84.9  & 87.1          & 82.5          & 84.7   & 86.9          & 80.2          & 83.4   \\
    ContourNet~\cite{ContourNet} & 87.6 & 86.1 & 86.9 &-&-&-& 86.9          & 83.9          & 85.4   & 84.1          & \textbf{83.7}          & 83.9  \\
    DRRG~\cite{RRG-Net} & 88.5 & 84.7 & 86.6  & 88.1 & 82.3 & 85.1 & 86.5          & 84.9            & 85.7  & 85.9          & 83.0            & 84.5  \\
    % GNNets~\cite{GNN} & 90.4 & 86.7 & 88.5 &-&-&-&-&-&- &-&-&-\\
    MOST~\cite{MOST} & 89.1 & \textbf{87.3} & 88.2 & 90.4 & 82.7 & 86.4 &-&-&-&-&-&-\\
    Raisi~\emph{et al.}~\cite{raisi2021transformer} & 89.8 & 78.3 & 83.7 & 90.9 & 83.8 & 87.2 &-&-&-&-&-&- \\
    TextBPN~\cite{TextBPN} &-&-&-&86.6&\textbf{84.5}&85.6&90.7&85.2&87.9&86.5&83.6&85.0 \\
    DBNet++~\cite{dbpp} &90.9&83.9&87.3&91.5&83.3&87.2&88.9&83.2&86.0&87.9&82.8&85.3 \\
    \midrule
    \textbf{DPTNet-Tiny(Ours)} & 90.3 & 77.4 & 83.3 & 88.4 & 82.9 & 85.6 & 88.9 & 83.9 &86.3 & 86.4  & 81.5	& 84.1 \\
    \textbf{DPTNet-Normal(Ours)} & \textbf{92.0} & 85.0 & \textbf{88.4} & \textbf{92.5} & 84.3 & \textbf{88.2} & \textbf{91.2} & \textbf{86.2} &\textbf{88.6}& 88.1  & 83.1	& \textbf{85.5} \\
    
    \bottomrule

    \end{tabularx}
    % \label{preform_compare1}
    \label{tab:compare}
\end{table*}

\begin{figure}[tbp]
\centering
\includegraphics[width=1.0\linewidth]{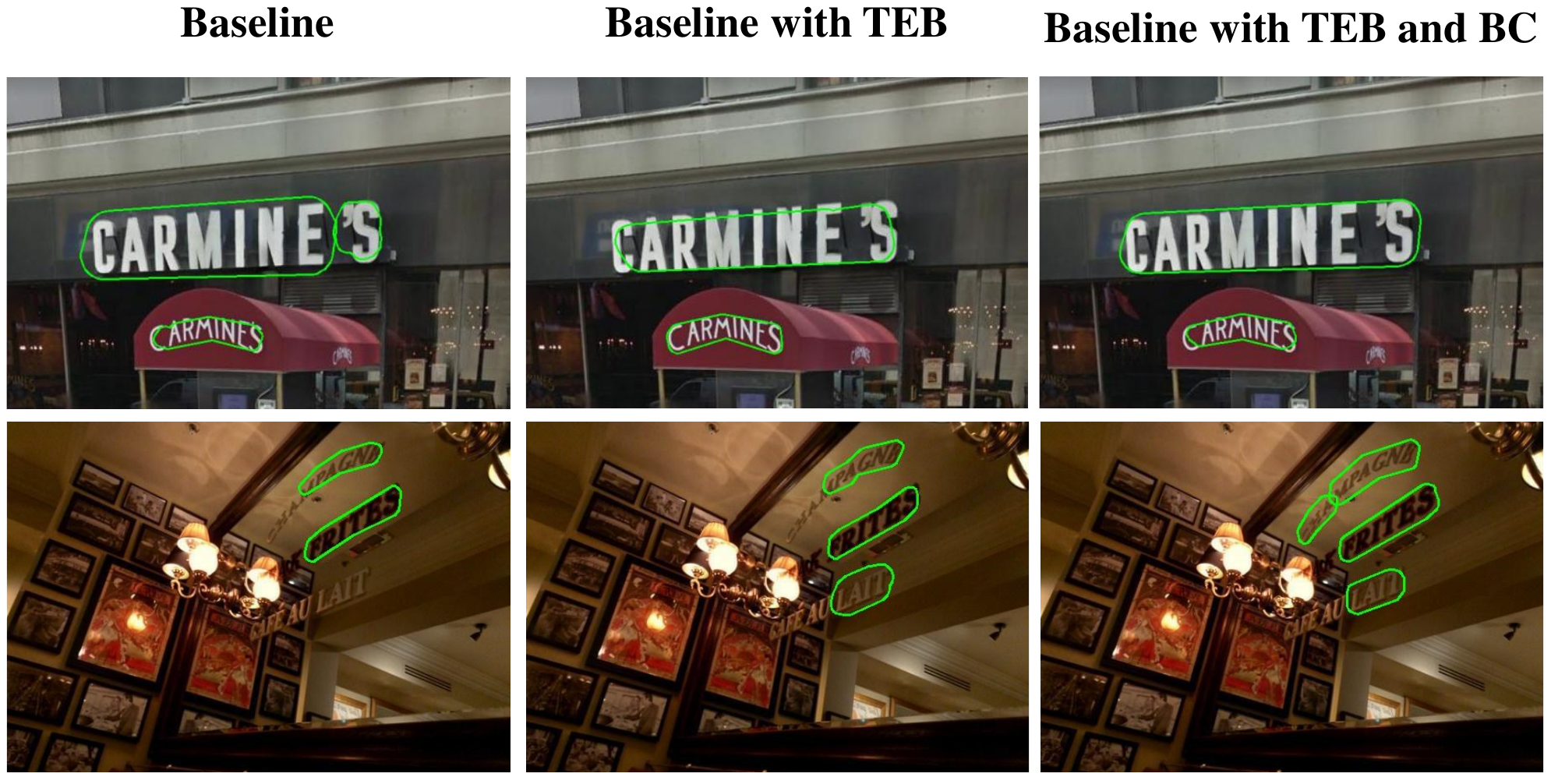}
\caption{Qualitative results w / o TEB and BC modules. Pictures are sampled from Total-Text.}
\label{visual}
\end{figure}

\subsection{Ablation Study}

Compared with previous segmentation methods, we introduce two main modules to improve the performance of text detection in our proposed DPTNet. An ablation study is conducted on the CTW1500 dataset and the MSRA-TD500 dataset. When the Transformer-enhanced backbone is not used, we replace it with standard ResNet at the corresponding size. The details of our designed experiments are shown in Tab. \ref{teb}.

{\bfseries Transformer-enhanced Backbone.}
As is shown in Tab. \ref{teb}, we can see that our proposed Transformer-enhanced backbone improves the performance significantly on both CTW1500 and MSRA-TD500. For light backbones, transformer-enhanced modules achieve 2.0\% performance gain on MSRA-TD500 and 1.4\% on CTW1500 respectively, in terms of F-score. For more complicated backbones, transformer-enhanced backbones bring 2.4\% improvements on MSRA-TD500 and 1.6\% on CTW1500, respectively. These results also show that more stacking blocks bring better performance while running slower. The DPTNet-normal outperforms the DPTNet-light by 2.1\% on MSRATD500 and 2.4\% on CTW1500, respectively, but costs twice the time of the light backbone.

% For the light backbone, Transformer-enhanced module achieves 2.0\% and 1.4\% performance gains in terms of F-measure on the MSRA-TD500 dataset and the CTW1500 dataset. For more complicated backbone, Transformer-enhanced backbone brings 2.4\% (on the MSRA-TD500 dataset) and 1.6\% (on the CTW1500 dataset) improvements. These results also show that the design with more stacking blocks achieves better performance than the lighter one but runs slower. The DPTNet-normal outperforms the DPTNet-light by 2.1\% (on the MSRATD500 dataset) and 2.4\% (on the CTW1500 dataset), but cost twice the time as the light backbone.

{\bfseries Bi-directional Connection Module.}
Ablation experiment results of the BC / TEB modules on MSRA-TD500 and CTW-1500 are shown in Tab. \ref{teb}. By adding a BC module to ResNets, we exchange feature information between two parallel convolutional layers, and this brings improvements on F1-score while reducing the inference speed. Moreover, the combination of the BC and TBE modules further improves the performance and rises F1-score by 2.8\% and 3.1\% over the baseline on both light and normal models, respectively.

% Tab. \ref{teb} shows the results with and without the proposed interaction module. The ablation experiments on MSRA-TD500 and CTW-1500 show that when the bi-directional connection module is not included, performance on standard ResNet backbone and proposed Transformer enhanced backbone deteriorates. For the experimental design of adding bi-directional connection module to ResNet, we exchange feature information between two parallel ResNet layers. Experiments show that the interaction between the two CNN paths does not greatly improve the overall effect, but reduces the processing speed. Moreover, the combination of Transformer-enhanced backbone and bi-directional connection module further improves the performance and rises the F-score by 2.8\% and 3.1\% beyond the baseline on both light model and normal model.

% As far as the F1-measure is concerned, without a bi-directional connection module, ResNet-18 drops 0.5\% and 1.1\% respectively on MSRA-TD500 and CTW-1500, DPTNet-tiny drop 1.1\% and 0.7\%, ResNet-50 drop 1.6\% and 0.2\%, DPTNet-normal drop 0.5\% and 0.5\%. 

% Obviously, with the support of the multi-head attention decoder, the performance of the model has made a further leap. Such a 

These results validate the effectiveness of our design. DPTNet can obtain richer fused features and more discriminative representations which benefit the text detection task by enlarging the receptive field. We visualize our detection results in Figure \ref{visual} for further inspection.

% the gradual improvement of 

% Figure 4: Qualitative results w / o TEB and BC modules. Pictures are sampled from Total-Text.

\subsection{Comparisons with Previous Methods}

\begin{figure*}[!tbp]
\centering
\includegraphics[width=0.9\linewidth]{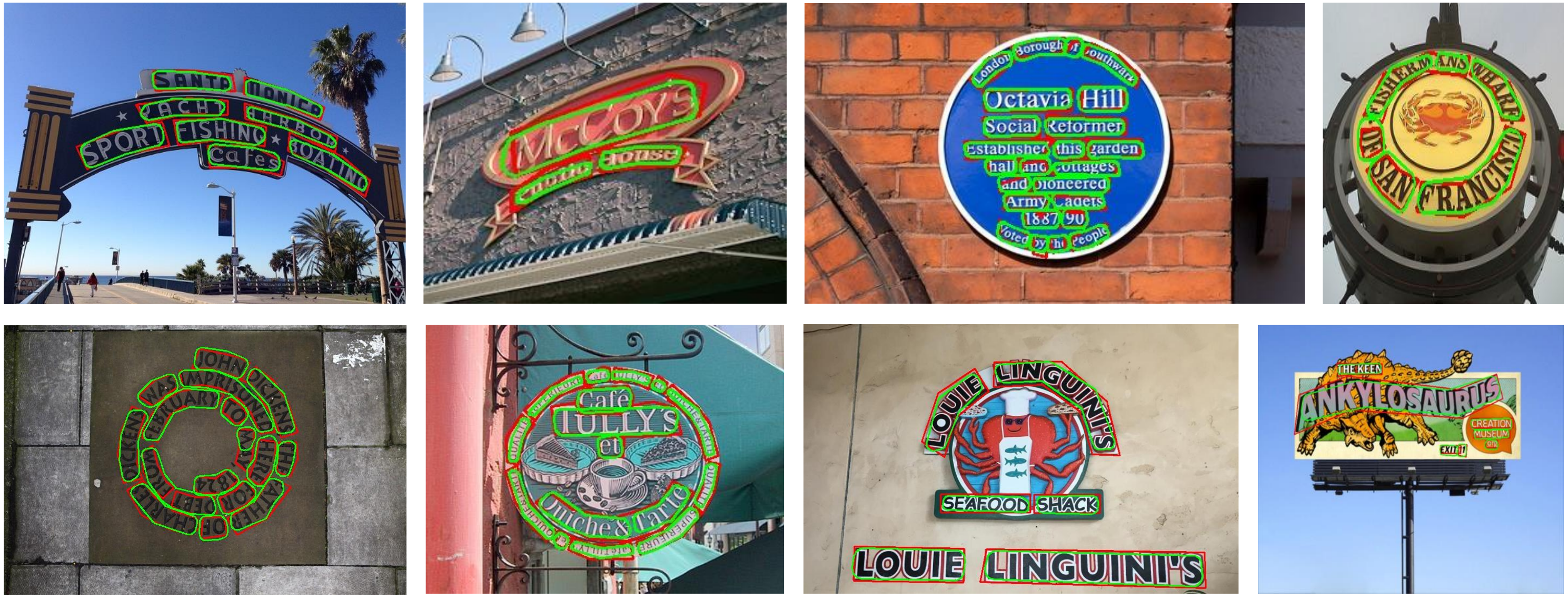}
\caption{Qualitative results obtained by our method. These pictures are sampled from Total-Text. We mark the ground-truth annotations with red and our detection results are in green.}
\label{excellent_res}
\end{figure*}

We evaluate our DPTNet on various datasets which contain challenging rotated, curved and long line-level instances, and compare it with the other competing works.

{\bfseries Curved text detection.}
We prove the shape robustness of our method on Total-Text, which has plenty of curved text instances. As shown in Tab. \ref{tab:compare}, our method achieves SOTA performance in both F1-score and inference speed. The proposed DPTNet-normal model outperforms the previous SOTA methods by 0.2\%, 1.0\%, 0.7\% and 0.2\% on ICDAR2015, MSRA-TD500, Total-Text and CTW1500, respectively. And the inference speed of the lighter model can be further accelerated with a small acceptable performance drop. This demonstrates the robustness of the model and shows the trade-off between speed and F1-score.

% As shown in Tab. \ref{tab:compare}, our method achieves SOTA performance on both accuracy and speed. The DPTNet-normal model outperforms the previous SOTA methods on their corresponding SOTA datasets (e.g. ICDAR2015, MSRA-TD500, Total-Text, CTW1500) by 0.2\%, 1.0\%, 0.7\%, 0.2\% respectively, and the test speed of the lighter model can be greatly improved with a small acceptable performance drop. This demonstrates the robustness of the model and shows the trade-off between speed and accuracy. 

{\bfseries Multi-oriented text detection.}
As introduced above, the ICDAR 2015 dataset is a multi-oriented text dataset. The difficulties for the detection task on this dataset are lots of small and low-resolution text instances. In Tab. \ref{tab:compare}, compared with DBNet \cite{MinghuiLiao2019RealtimeST}，our DPTNet outperforms 0.9\% in F1-score while keeping a similar speed. Compared with the former DETR-based method~\cite{raisi2021transformer}, our proposed model shows a much better detection performance (88.4\% \emph{vs.} 83.7\%) on the small and blurry text of ICDAR2015. Obviously, DPTNet is superior to previous methods in F1-score and speed.

% compared with DB \cite{MinghuiLiao2019RealtimeST}, under the condition that the speed is basically similar, the F1-score has been raised to a new height. 

{\bfseries Comparisons with Transformer-based backbone.}
We perform the pre-training job on the SynthText dataset for 160k iterations. After the pre-training, we finetune the models for 1200 epochs on Total-Text. As shown in Tab. \ref{tab:params}, our method achieves the best trade-off among performance, model size, and computational cost. The DPTNet surpasses other Transformer-base backbones with a large margin of 2.8\% in F1-score. It is obvious that pure Transformer-based backbones are not as effective as our fused ones.

% Before training on Total-Text

% The tab. \ref{tab:params} shows our successful trade-off between the F1-score, model size and computation cost. It can be seen that we are absolutely lightweight in terms of overall computation. Our model size and computation cost are second to none, and surpass the other backbones with a large margin of 2.8\% in the F1-score. It is obvious that pure Transformer-based backbones are not as effective as our fused one. 

%  The arrows The best results are highlighted by bold.
\begin{table}[ht]
\centering
\caption{Comparisons of detection results on Total-Text with prevailing Transformer-based backbones. ``Params" refers to the quantity of parameters. ``FLOPs" is an index calculated under the input scale of $224 \times 224$. F1 is in $\%$.}
\begin{tabularx}{1.0\linewidth}{@{}l*{3}X@{}}
\toprule
Method        & ${\rm Params_{\downarrow}}$    & ${\rm FLOPs_{\downarrow}}$     & ${\rm F1_{\uparrow}}$\\ \midrule
Swin-T \cite{swin-transformer}          &29M	&4.5G	&79.8 \\
Swin-S \cite{swin-transformer}         &50M	&8.7G	&81.8 \\
CvT-13 \cite{cvt}         &20M	&4.5G	&82.1 \\
CvT-21 \cite{cvt}         &32M	&7.1G	&83.2 \\
CMT-S \cite{cmt}         &25M	&4.0G	&82.9 \\
CMT-B \cite{cmt}         &46M	&9.3G	&84.1 \\
Conformer-S \cite{conformer}          &26M	&3.6G	&84.8 \\
Conformer-B \cite{conformer}         &38M	&10.6G	&85.8 \\
\midrule  
DPTNet-Tiny(ours)      & \textbf{11M} & \textbf{1.2G} & 86.3 \\ 
DPTNet-normal(ours)    & 19M & 3.1G & \textbf{88.6} \\ 

\bottomrule
\end{tabularx}
\label{tab:params}
\end{table}

\begin{table}[!ht]
\centering
\caption{Detection results (in \%) on Total-Text and TD500. The values in the table mean the F1-score of each detector on the corresponding dataset. The best results are highlighted in bold.}
\begin{tabularx}{1.0\linewidth}{@{}l*{2}X@{}}
\toprule
% \begin{tabular}{|c|c|c|c|c|c|c|c|c|c|c|}
% 
% \multirow{2}{*}{Backbone} & \multirow{2}{*}{BC}  & \multicolumn{4}{c}{MSRA-TD500} & \multicolumn{4}{c}{CTW1500} \\ \cline{3-10} 
methods/datasets      & Total-Text      & TD500\\ 
\midrule   
Raisi et al. \cite{raisi2021transformer}  & - & 87.2 \\ 
FSG \cite{tang2022few}           &  88.1  & 87.9 \\
ASTDT \cite{raisi2022arbitrary}        &  87.8  & 86.5 \\
SwinText \cite{huang2022swintextspotter}         &  88.0  & - \\
TESTR \cite{textspotting}         &  88.0  & - \\
\midrule  
% DPTNet-normal                 &  $\times$            &  91.9   & 83.1   & 87.2   & \textbf{29}  & 87.3 & 81.6 & 84.4 & \textbf{21} \\ 
DPTNet-normal &  \textbf{88.6}  & \textbf{88.2} \\
\bottomrule
\end{tabularx}
\label{tab:compareTran}
\end{table}

{\bfseries Comparisons of models combining CNN with Transformer.}
There are other works besides ours trying to combine CNN with Transformer and apply it to text feature extraction. These methods mainly connect the CNN and Transformer modules in series in an end-to-end fashion. As can be seen from tab. \ref{tab:compareTran}, our model achieves state-of-the-art performance on both Total-Text and TD500 datasets. We outperform the best-performing FSG model on the Total-Text and TD500 datasets by 0.5\% and 0.3\%, respectively. We visualize our detection results in Figure \ref{excellent_res} for further inspection.

\section{Conclusion} \label{conclusion}

In this paper, we proposed a dual-path network named DPTNet to alleviate restrictive receptive fields and the weak modeling capability of the original CNN module for scene text detection. With the proposed bi-directional interactions across two parallel branches, DPTNet provides complementary clues in the local and global views. The learned features can thereafter be used to identify the segmentation masks for improving their robustness. The extensive experiments on four challenging benchmarks demonstrate the effectiveness and generalization ability of our DPTNet. We would hope that our method can serve as a solid baseline for segmentation-based methods on scene text detection and also motivate further research. 

\newpage

% Use \bibliography{yourbibfile} instead or the References section will not appear in your paper
\nobibliography{aaai23}
% \bibliography{aaai23}

\end{document}